%% file: main.tex
\documentclass[11pt]{article}

\usepackage[]{naacl2021}

\usepackage{times}
\usepackage{latexsym}
\usepackage{times}  
\usepackage{helvet} 
\usepackage{courier}  
\usepackage{graphicx} 
\graphicspath{ {./} }
\usepackage{bbm}
\usepackage{todonotes}
\usepackage{amsmath}

\usepackage[utf8]{inputenc}

\usepackage{microtype}

\title{
Picking Pearl From Seabed: Extracting Artefacts from Noisy Issue Triaging Collaborative Conversations for Hybrid Cloud Services
  \author{
  Amar Prakash Azad$\dagger$, \hspace{5mm} {Supriyo Ghosh$\ddagger$ }, \hspace{3mm} Ajay Gupta $\dagger$, \\
  amarazad@in.ibm.com, \hspace{3mm} supriyog@ibm.com,  \hspace{3mm}ajaygupta@in.ibm.com\\
   {\bf Harshit Kumar$\dagger$},  \hspace{3mm}  {\bf Prateeti Mohapatra$\dagger$} \\
     harshitk@in.ibm.com,  pramoh@in.ibm.com\\
     $\dagger$ IBM Research Lab, India\\
     $\ddagger$ IBM Research Lab, Singapore }
}

\begin{document}
\maketitle
\begin{abstract}
Site Reliability Engineers (SREs) play a key role in issue identification and resolution. After an issue is reported, SREs come together in a virtual room (collaboration platform) to triage the issue.
While doing so, they leave behind a wealth of information which can be used later for triaging similar issues. 
However, usability of the conversations offer challenges due to them being i) noisy and ii) unlabelled.  
This paper presents a novel approach for issue artefact extraction from the noisy conversations with minimal labelled data. We propose a combination of unsupervised and supervised model with minimum human intervention that leverages domain knowledge to predict artefacts for a small amount of conversation data and use that for fine-tuning an already pre-trained language model for artefact prediction on a large amount of conversation data.
Experimental results on our dataset show that the proposed ensemble of unsupervised and supervised model is better than using either one of them individually.

\end{abstract}
\section{Introduction} 
\begin{figure}[!tb]
  \includegraphics[width=\linewidth, height = 0.45\linewidth]{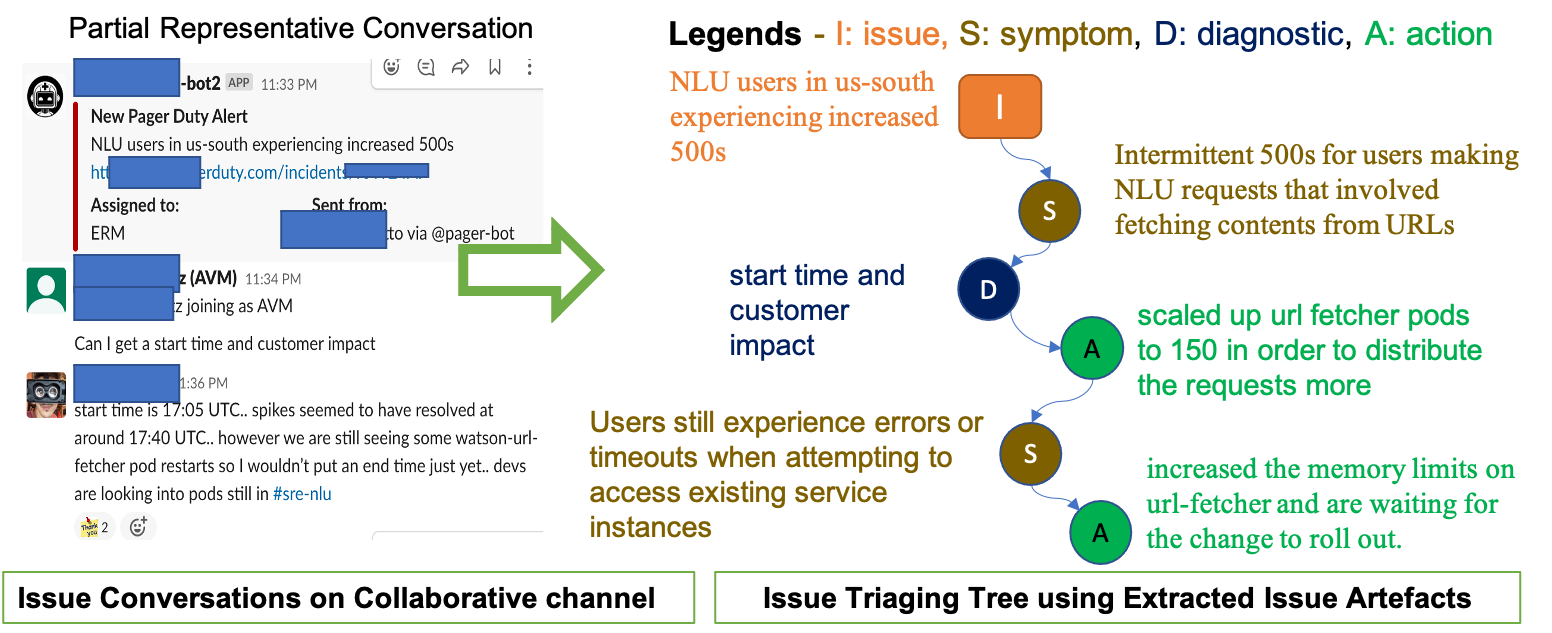}
  \vspace{-0.7cm}
  \caption{\small {An Issue Triaging Tree example from Collaborative Channel Issue conversations (sensitive information blurred)}. }
  \label{fig:issueTT}
  \vspace{-0.6cm}
\end{figure}

The deployment of applications using the micro-services architecture has simplified the scope of developers to put the system in production, however, the role of Site Reliability Engineers (SREs) 
has become more complex. Most times, when services fail leading to alerts and anomalies, a Site Reliability Engineer (SRE) comes into play whose role is to ensure that the services run uninterrupted, i.e. if they fail, they return to normal execution as quickly as possible without impacting any client business. 
Collaboration platforms such as Slack, Microsoft Teams in an IT operation management act as a virtual war room that allows teams to collaborate and engage in conversation with each other, with a common aim to identify the problem,  symptoms, diagnosis and resolution to resolve the issue. 
Such conversations contain useful artefacts including symptoms, diagnosis and action (key information of SRE’s interest) for the issue. These artefacts can be used to find similar conversations from the historical conversations database - this requires grounding historical conversations and the current conversation to a common skeleton structure, defined as triaging tree. Triaging tree is a structured temporal view of key artefacts pertaining to an issue. It provide SREs a holistic view of an issue and its remedial steps in a consolidated format which is easy to consume and interpret. 
Figure~\ref{fig:issueTT} depicts an example where key artefacts from noisy SREs' \emph{Conversation} (with chit-chat) are extracted and converted to an \textit{Issue Triaging Tree}. 

Building a \textit{triaging tree} for a conversation requires utterances labeling with artefacts. 
Labeled data scarcity is one of the major bottleneck to build an AI system, same is the case here; we do have a lot of conversation data, however, artefacts annotations are not available for the utterances in these conversations. To overcome this problem, we annotated a small conversation dataset using manually defined dictionaries and rules. Further, we get this small labeled data corrected by SREs such that we have a clean labeled data for the next step which involves fine-tuning a pre-trained language model such as BERT for the artefact labeling task. The motivation behind manually correction step is to ensure that utterance labeled with artefacts, albeit small, should be 100\% accurate, such that the underlying language model that would be fine-tuned on this data for artefact prediction is free of any human biases. 



In this paper, we propose a novel approach for artefact prediction from conversations through an ensemble of  i) Domain Knowledge Guided unsupervised artefact prediction model  ii) A supervised approach with BERT model fused with FastText embedding. More particularly, the proposed approach outperforms the artefact prediction in absence of label data. With minimal label verification effort, it achieves improved performance which outperforms existing supervised models.  

\section{Proposed Method}
\begin{figure}[!tb]
  \centering

  \includegraphics[width=\linewidth, height = 0.4\linewidth]{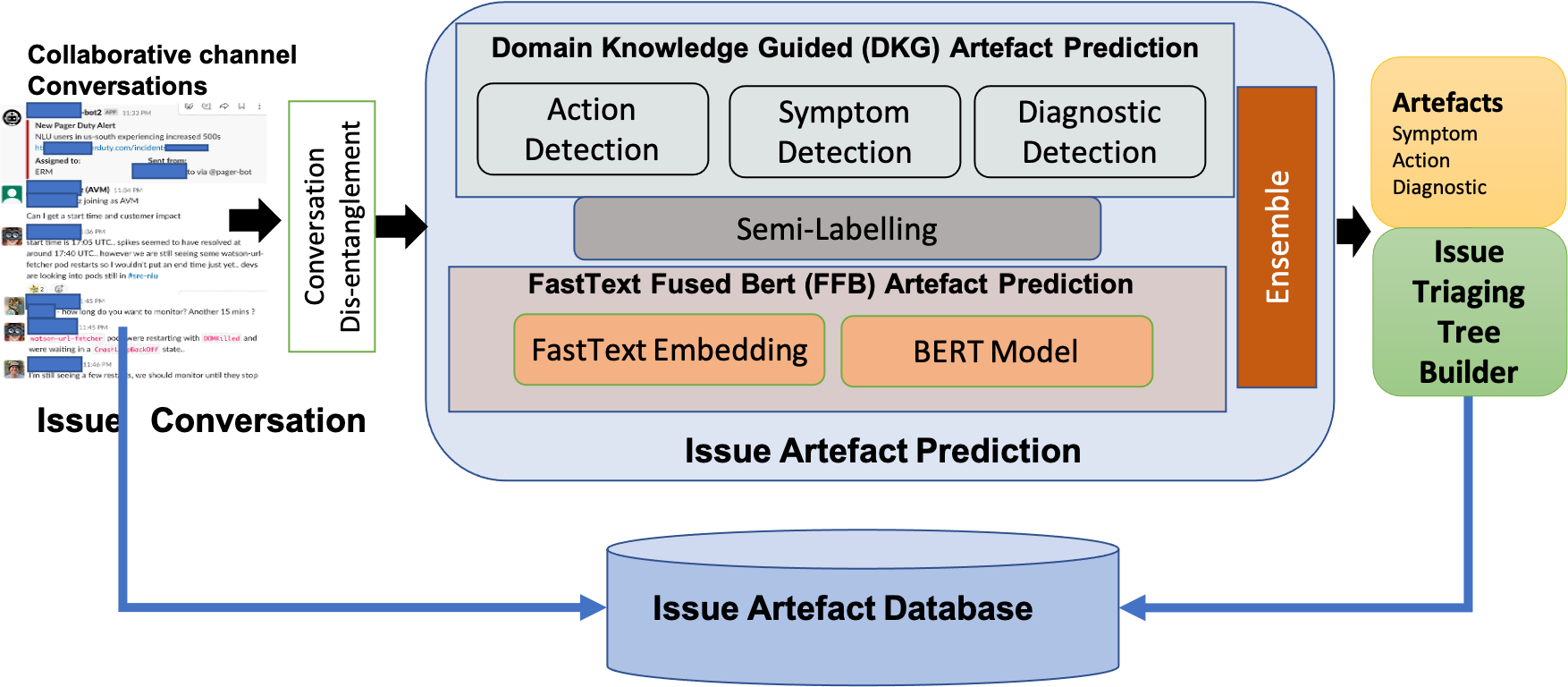}
 \caption{Proposed framework for Artefacts Prediction from Issue Conversations}
  \label{fig:method}
 \vspace{-0.5 cm}
\end{figure}

The proposed framework, depicted in Figure~\ref{fig:method}, consists of the following modules. The \textit{Conversation Disentanglement} module separates intermingled multi-interlocutor utterances into coherent conversations with clear start and stop boundaries. The \textit{ Artefact Prediction} module labels each utterance in an issue conversation with one of the three artefacts (Symptom, Action, and Dignostic) and chit-chat(CC) using a combination of unsupervised and supervised learning approaches. The input to the Triaging-Tree-Builder module is a labelled conversation with artefacts, and the output is a triaging tree which is stored in an Issue Artefact Database for later consumption to generate an issue summary or finding similar issues. 

\subsection{Conversation Disentanglement}

\input{conv_disent.tex}
\subsection{Artefact Prediction
}
\label{s:artefact}
After obtaining disentangled conversations, the next step is to automatically label utterances in a conversation with artefacts. Each conversation contains a large number of chit-chat utterances besides the utterances that are helpful for issue resolution. 
The {\it Artefact Prediction} module assigns relevant artefacts to utterances, in the process, labeling all irrelevant utterances with artefact chit-chat. We formulate the problem as a multi-class classification problem which includes the artefact label set as, Symptom, Action, Diagnostic and Chit-chat. 
The first step is an unsupervised approach, a \textit{ Domain Knowledge Guided} (DKG) artefact prediction approach that uses a set of predefined rules to predict artefact for each utterance in a conversation. In the second step, the labelled utterances are validated by a human for its correctness. The final step is a supervised learning model, \textit{FastText Fused BERT} (FFB), that uses the labeled conversations as training data to further label the rest of the conversations. We improve the artefact prediction model by taking an ensemble of DKG and FFB, thereby getting the best of both worlds. 


\subsection{ Domain Knowledge Guided (DKG) Artefact Prediction}
\label{sec:unsup_id}

The Domain Knowledge Guided (DKG) artefact prediction module consists of three sub-modules, each of which aims to detect utterances of an artefact type, namely, Action, Symptom, Diagnostic in an unsupervised manner. Remaining utterances are labelled as chit-chat.

 \subsubsection{Action Artefact Detection}
 Action detection is based on the observation that verb-noun pair usually provides natural and meaningful clusters of IT tickets  ~\cite{Ayachit20}.  For a given utterance, key entities are extracted and linked to appropriate actions. 
The approach for action utterance detection consists of three steps: (i) candidate action utterance selection using action verbs from the domain-specific dictionary, ii) extraction of key entities in these utterances, and (iii) link key entities and action verb using Semantic Role Labeling  \cite{roth2014composition} (shallow semantic parser) to filter valid action utterances. 

In IT domain, an \textit{action} is defined as a process of performing a change operation by engineers to fix an issue. In particular, action words are those verbs which results in state change of an entity, e.g. ~\textit{increase}, \textit{reboot}. We curated \textit{action word} dictionary using existing Technical Support and Operations corpus which consists of changes and service request documents. We perform candidate utterance selection in presence of an \textit{action word}. 

Extraction of key entities from utterances is inspired by the approach in~\citet{mohapatra2018domain}. The approach uses both linguistic and non-linguistic features to determine an entity. In step (iii), we use Semantic Role Labelling(SRL) ~\cite{pradhan2004shallow, Gildea} to extract action-entity links from utterances. SRL is a shallow semantic parsing task which provides answers of who did what to whom, when, etc. 

\begin{figure}
  \includegraphics[width=\linewidth]{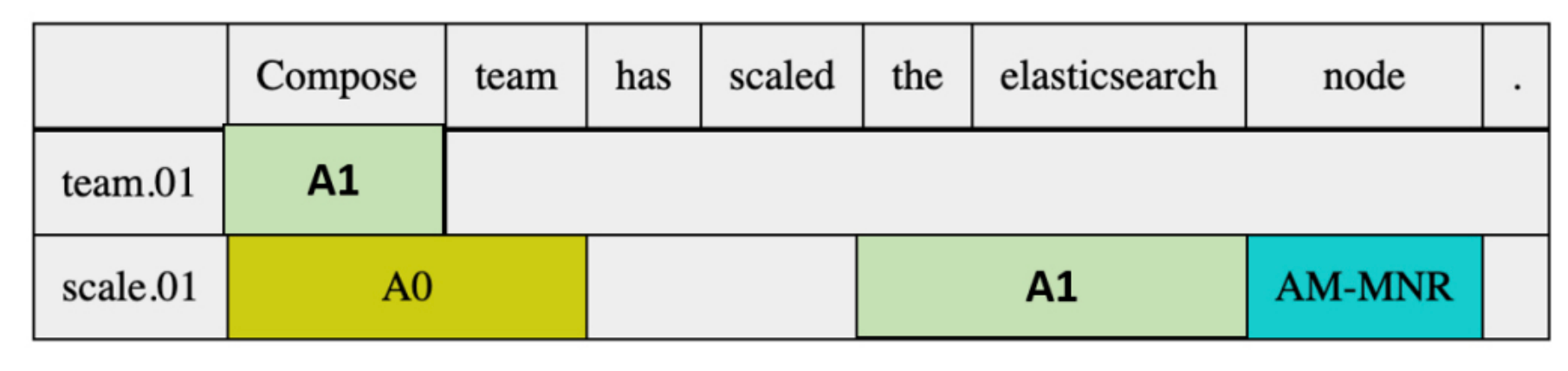}
  \caption{Example of Semantic Role Labeling}
  \label{fig:srl}
  \vspace{-0.5cm}
\end{figure}

We show in Figure~\ref{fig:srl} 
the semantic roles with numbered arguments and adjuncts for a sample utterance. 
Each row in the figure  depicts the label of an argument with respect to a particular predicate. In the example sentence, \textit{``team'', ``scale''} are the two predicates. We use a predicate in a sentence, if it's part-of-speech is Verb and if it is presented in the action dictionary. We also explored the relation between semantic role types and ground truth key phrases annotated for each technical document in ~\cite{mohapatra2018domain}, by deriving the distribution for each role type of a key phrase. We obtained the semantic role of each word in the ground truth key phrase from the corresponding sentence to identify the overall frequency distribution of each role in the corresponding dataset. From this experiment, we found that the semantic role \textit{A1} was the most dominating role. Hence, for each predicate in a sentence, we used only the corresponding text that had \textit{A1} as its semantic role. 
We extract action utterance if positive action-entity links are present in the utterance. 
\subsubsection{Symptom Artefact Detection}
\label{s:symptom}
In the support domain, \textit{symptom} artefacts are typically governed by the symptom key-terms, e.g. 500x errors, failure, etc.  We curated a \textit{symptom word} dictionary using existing Technical Support and Operations corpus which consists of changes and service request documents.
Guided by the presence of dictionary terms in an utterance helps in identification of utterances of artefact type Symptom, and the phrase surrounding the matching term is the actual artefact for the symptom.

\subsubsection{Diagnostic Artefact Detection }
\label{s:Dignostic}
Diagnostic artefact detection is an important component of information mining for automated incident remediation. Knowing the diagnosis-related utterances can help to understand what investigations were carried out to resolve an issue. In the conversations, we noted that most of the diagnostic utterances are questions or query type statements and they are either action type or symptom type utterances. For diagnostic artefact extraction, we identify query or question utterance from the action and symptom utterances.  
Query utterances, can take both explicit and implicit lexical forms \cite{Mckewon2004,Forsythand2007},
 e.g., explicit: \textit{``Which services are affected ?"}, implicit: \textit{``I was wondering what is the latest impact."}. Furthermore, the queries may also contain informal utterance construct as the conversations are informal in nature. 
To identify the queries, we adopted a semi-supervised approach augmented with lexical rules along with a simple and effective Naive Bayes classifier. The lexical rules are apt to capture queries containing question words (mainly constituting 5W1H question words, for e.g. who, when, where, what, why, how and presence of '?'), along with a set of other curated verb and adverb based question words (e.g., could, kindly, please). To detect implicit queries which capture the informal query utterances, we trained a Naive Bayes model on NPS Chat dataset\footnote{http://faculty.nps.edu/cmartell/NPSChat.htm}.
When both lexical and Naive Bayes model yield a negative label for an utterance, then it is labeled as a negative query.
Since the artefact detection modules are independent of each other, an utterance can be tagged to multiple artefact types. The Diagnostic artefact has the highest priority followed by Action and Symptom artefact classes. 

\subsection {Semi-Labelling}
The unsupervised artefact detection from issue conversations described above doesn't require any label data, thus addressing the problem of cold start for artefact detection. We use the module to obtain conversation utterance pre-label. We ask SRE's to verify and correct the pre-labels which can be used for the following supervised artefact detection model. One of the reasons to provide these pre-label is to minimize the human cognition effort and time as SRE's are the domain expert but have serious time scarcity for complete label annotation.

For the label correction, 350 utterances were distributed among four SRE's with overlapping issue conversations. We observed reasonably good agreement among SRE's and assign final label to an utterance based on majority voting. Note, due to time scarcity of SRE's the number of annotations are not in abundance as in typical supervised classification approaches. 

\subsection{FastText Fused with BERT (FFB) Artefact Prediction}
The section explains the supervised approach for artefact prediction using the labelled data obtained from the previous step, i.e. semi-labelling method. 
Despite the fact that fine-tuning a pre-trained BERT model\cite{devlin2018bert}  can produce attractive performance for classification task on natural language understanding tasks\cite{chen2019bert}, such pre-trained BERT model fails to reproduce the performance on our artefact detection task due to limited training data and complexities associated with identifying the right representation of technical terms that are unique to our problem domain. Unfortunately, training a domain-dependent BERT model from scratch is infeasible for us as it is computationally expensive and requires a huge amount of relevant training data. Therefore, we employ a pre-trained BERT model to capture the understanding of natural language conversation and augment its capability with a light-weighted domain-dependent trained model. To obtain a light-weighted model that can represent the unique technical terms appropriately, we train a FastText model \cite{bojanowski2017enriching} using our technical domain dataset.

\begin{figure}[!htb]
	\centering
	\includegraphics[width=0.42\textwidth, height = 0.5\linewidth]{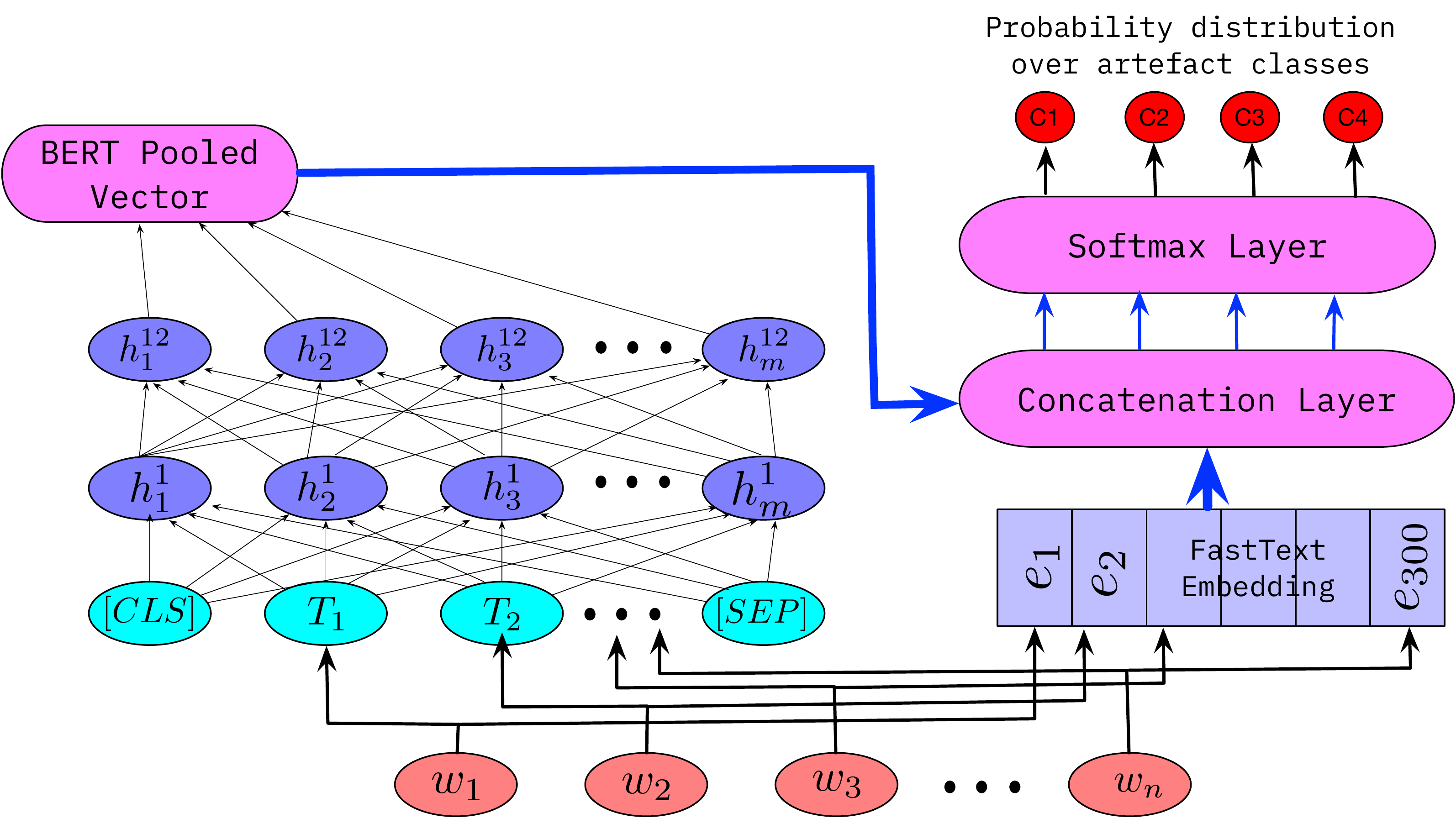}
	\caption{ 
	Architecture of FastText Fused BERT model.}
	\label{fig:supervisedArchitecture}
	\vspace{-0.2cm}
\end{figure}

Figure \ref{fig:supervisedArchitecture} delineates the architecture of our proposed FastText fused BERT (FFB) based artefact prediction method. 
We employ a 12-layer pre-trained BERT model\footnote{ https://tfhub.dev/tensorflow/bert\_en\_uncased\_L-12\_H-768\_A-12/1 .}, which is a multi-layer bidirectional transformer encoder built upon the transformer model proposed by \cite{vaswani2017attention}. 
Given an input utterance, we begin by tokenizing the utterance and insert a special [CLS] token at the beginning and a special [SEP] token as the final token. Then, a concatenation of \emph{WordPiece} embeddings, positional embeddings, and the segment embedding of the utterance tokens are fed to the BERT model as an input representation.
Let the output of the BERT be $\boldsymbol{H}=(\boldsymbol{h_1},\boldsymbol{h_2},...,\boldsymbol{h_N})$ for a given input $\boldsymbol{x}=(x_1,x_2,...,x_N)$. The hidden representation of the [CLS] token, denoted by $\boldsymbol{h_1}$, is used for artefact prediction, as it provides a sentence level representation of the input text. In addition, we pass the input utterance to a domain-specific trained FastText model that produces a 300-dimensional embedding of the utterance, $\boldsymbol{e}.$ 
Finally, we concatenate the [CLS] token output from BERT model and the FastText output embedding vector to obtain the final hidden layer output, $\boldsymbol{f} = (\boldsymbol{h_1} + \boldsymbol{e}),$  and pass it through a \emph{softmax} layer to get the probability distribution over the artefact classes which is expressed as   $ \boldsymbol{y}^i = \text{softmax}(\boldsymbol{W}^i \boldsymbol{f}+\boldsymbol{b}^i) \label{eq:consoftmax}.$


\subsection{Ensemble Method for Artefact Prediction}
\begin{figure}[!htb]
    \small
	\centering
	\includegraphics[width=0.45\textwidth, height = 0.2\linewidth]{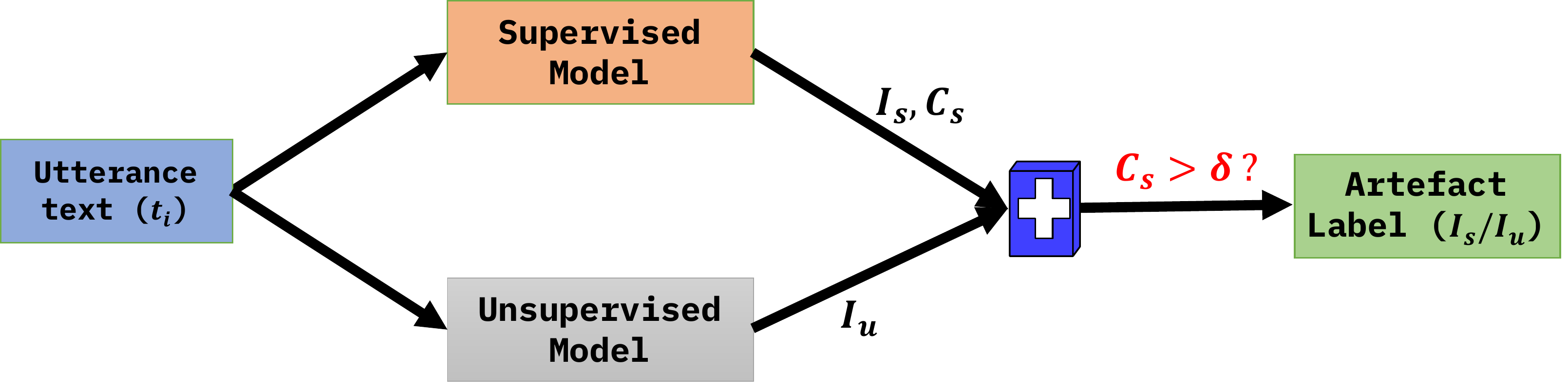}
	\caption{
	Ensemble method Architecture}
	\label{fig:ensembleArchitecture}
	\vspace{-0.8cm}
\end{figure}

We experimentally observed that the performance of our supervised approach is not always superior than the unsupervised approach. This could be due to limited amount of labelled data and complexities associated with technical terms, the confidence score of the supervised artefact classification algorithm is relatively low in many cases. 
Therefore, we propose a simple and effective ensemble method to combine the power of both the supervised and the unsupervised artefact detection approaches. The key idea behind our ensemble approach is that we only rely upon the classification results of the supervised approach if the confidence score is high; otherwise we resort back to the decision of the unsupervised approach. Figure~\ref{fig:ensembleArchitecture} illustrates the architecture of our proposed ensemble method, where the final artefact class label is taken from the decision of the supervised model, only if the confidence score is higher than a threshold parameter $\delta$\footnote{We set the value of $\delta$ to 0.9 in our experiments, which is obtained using grid search.}, else the class label from the unsupervised approach is considered as the final label. 

\subsection{ Building Issue Triaging Tree }
After artefacts are predicted for an issue conversation, we build \textit{issue triaging tree}.  
As a naive implementation, we create a connected tree of artefacts in order of their temporal sequence as shown in Figure \ref{fig:issueTT} where the nodes represents artefact types. 
The issue triaging tree presents a concise structured representation of issue remedial or triaging. 
\section{Results and Analysis}
This section evaluates our method on two dimensions: (i) Efficacy of conversation disentanglement module in terms of identifying segmented conversations; and (ii) Accuracy of the detected and extracted ``artefacts" from the segmented conversations.
\subsection{Conversation Disentanglement}
To evaluate the conversation disentanglement module, we extracted threads and contextual messages from conversation data. We extracted $189$ threads where two consecutive threads are separated by at least $24$ hours to avoid overlap. 
For each thread, we extracted up-to $50$ contextual messages before the first message and after the last message. These contextual messages are extracted within a two hours window. We extracted $508$ contextual messages for our experiments. For evaluation, we manually annotated contextual messages to identify if messages are part of the thread or not.  The precision and recall of the module are $67\%$ and $83\%$ respectively. Table \ref{tab:conv_seg} provides the key performance statistics of our disentanglement module. We observe that most of the false positive cases resulted because users who are participants in thread messages are also a participant of contextual messages. Most of these messages are of type \textit{change request access} (e.g., \textit{@bot Please provide access to $\langle user \rangle$ to cluster}). 
\begin{table}[h]
    \small
    \centering
    \begin{tabular}{|c|c|c|c|c|c|}
   \hline
     \# Conversations & \# Messages & TP & FP & TN & FN  \\ \hline
      189  & 508 & 58 & 29 & 409 & 12 \\ \hline
    \end{tabular}
    \caption{Performance analysis of the Conversation Disentanglement method on our dataset.}
    \label{tab:conv_seg}
    \vspace{-0.6cm}
\end{table}

%
%


\subsection{Artefact detection}

\noindent {\bf Baseline methods:} 
We employ two state-of-the-art baseline methods for performance comparison against our proposed artefact detection approaches: (i) K-Means clustering ~\cite{popov2019unsupervised} -- an unsupervised approach in which conversation utterances are clustered over TF-IDF vector space and utterance labels are assigned based on majority voting for each cluster. We use it as a baseline against our unsupervised artefact extraction approach (DKG); and (ii) BiLSTM-CRF model \cite{CRF-BILSTM} -- a supervised approach where utterance representations are learned using the BiLSTM model which are then used for predicting utterance labels modeled as a sequence labeling task. We use it as a baseline against our supervised FFB approach.
%


\noindent{\bf Dataset description:} 
For artefact detection evaluation, we randomly obtained $280$ issue conversations that contained $7975$ utterances from the collaborative channel involving hybrid cloud related operations and support work. 
Table \ref{tab:datastats} shows the artefact distribution statistics of the dataset. 
\begin{table}[!htb]
    \small
    \centering
    \begin{tabular}{|c|c|c|c|c|c|}
    \hline
    \# Utterances & Sym & Act & Diag &  Chit-Chat \\ \hline
       7975 & 18.4\%  & 3.2\%  & 20.1\% & 58.4\% \\  \hline
       \end{tabular}
       \vspace{-0.2cm}
    \caption{Data distribution over artefact types. }
    \label{tab:datastats}
    \vspace{-0.3cm}
\end{table}
For FFB approach, we used FastText embeddings trained on IT domain support data corpus consisting of documents (e.g., troubleshooting guides, manuals, etc) curated from Red Hat and other cloud services sites. Around $2$ million sentences were extracted from these documents. 
We observe that the FastText model, which was trained from scratch, provides us a better word relations compared to BERT Language models (uncased). 
We trained FastText for $300$ epochs to get a 300-dimensional vector representation for each word and considered sub-words from length $3$ to $20$. This enabled us to handle multiple word phrases like ``Red Hat".

\noindent {\bf Experimental Evaluation:}
For evaluation of our proposed artifact detection algorithms, we created a test set consisting of seven issue conversations with about $400$ utterances. We obtained annotations of the test set from four SREs to ensure inter-annotation agreement. 
 %
Considering SREs' time scarcity for annotation, we choose this relatively small sized test set. We used Precision, Recall and F1 measures to evaluate different artefact prediction methods.  
We compare the performance of DKG  with KMeans, FFB model with BiLSTM-CRF and the Ensemble model against both DKG and FBB. 
The performance evaluation on our test set is presented  in Table~\ref{tab:averageprecision} and Table \ref{tab:res}.


As shown in Table~\ref{tab:averageprecision}, the unsupervised artefact detection approaches using K-Means clustering and the proposed DKG method have an average precision of $36.7$ and $74.1$, respectively. 
This extraordinary performance improvement can be attributed to the manually curated action dictionary, symptom dictionary and lexical rules used for action artefact detection, symptom artefact detection and diagnostic artefact detection, respectively. 
The supervised artefact detection methods using our FBB model improves the average precision from $51.5$ to $69.2$, an improvement of 25.5\%, over the BiLSTM-CRF model.
The proposed ensemble method, that leverages both unsupervised and supervised approaches, outperforms all the approaches by further improving the average precision to $77.7$.
\begin{table}[!tb]
\small
\centering
\begin{tabular}{|l|c|}
\hline
    Method        & \textbf{Average Precision} \\ \hline
K-Means\cite{popov2019unsupervised}  & 36.7                       \\ 
Bi-LSTM-CRF\cite{CRF-BILSTM} &   51.5                  \\ 
DKG         & 74.1                       \\ 
FFB &69.2 \\
ENSEMBLE    & \textbf{77.7}     \\ \hline                 
\end{tabular}
\vspace{-0.3cm}
\caption{Performances  of artefact prediction methods}
\label{tab:averageprecision}
    \vspace{-0.1cm}
\end{table}

As shown in Table \ref{tab:res}, with the availability of labeled data, the proposed FFB approach provides better recall as well as precision in all cases. The is because the FFB approach captures both the sequential structure of sentences through BERT and the domain-specific semantic through FastText embedding trained on support documents. However, it is interesting to note that the DKG performs better than FFB approach in some cases (e.g., P@CC and Re@Sym). This can be attributed to the lack of sufficiently labelled data for FFB, which is often a major challenge in industrial setting. 
Finally, the Ensemble approach leverages the power of both unsupervised and supervised approaches by combining DKG and FFB, yielding the best performance for all the artefact type detection.

We observed that existing approaches fails to perform well due to noisy nature of semi-formal conversation. For example, the utterance ``That you are updating the case?'' is labelled \textit{diagnostic} where as it is a \textit{Chit-Chat}.  
Our proposed approaches DKG and FFB can handle such noisy utterances, as DKG uses curated dictionary and FFB uses embeddings trained on 
IT domain dataset. 
However,in some cases our model also faces challenges, e.g. compound utterances when multiple artefact types are present. For example, ``We are seeing significant improvement in service since 11:30 UTC and continue to work on restoring all operations'' is labelled as \textit{action} while the ground truth is \textit{symptom}. 

\begin{table}[!tb]
    \small
    \centering
    \begin{tabular}{|c|c|c|c|c|}
    \hline
      Algo & DKG & Bi-LSTM CRF& FFB& Ensemble  \\  \hline
       P@Sym  		& 0.77  & 0.48 & 0.85 & \textbf{0.81} \\
       P@Act  		& 0.60  & 0.2 & 0.48 &\textbf{ 0.51} \\
       P@Diag  		& 0.48  & 0.64 & 0.56 &\textbf{ 0.48} \\
       P@CC 	 & 0.78  & 0.55 & 0.69 & \textbf{0.84} \\\hline
       Re@Sym  	& 0.72  & 0.48 & 0.53 & \textbf{0.75} \\
       Re@Act 		& 0.52  & 0.58 & 0.68 & \textbf{0.64} \\
       Re@Diag  	& 0.28  & 0.12 & 0.32 & \textbf{0.38} \\
       Re@CC 		& 0.88  & 0.85 & 0.86 & \textbf{0.87} \\\hline
       F1@Sym   	& 0.75  & 0.48 & 0.65 &\textbf{ 0.78} \\
       F1@Act  		& 0.56  & 0.29 & 0.57 & \textbf{0.57} \\
       F1@Diag  	& 0.35  & 0.20 & 0.40 &\textbf{ 0.38} \\
       F1@CC  		& 0.82  & 0.66 & 0.77 & \textbf{0.85} \\\hline
    \end{tabular}
    \vspace{-0.3cm}
    \caption{Performance analysis for Artefact detection methods with Precision (P), Recall (Re) and F1 score  (F1) for each artefact
   . }
    \label{tab:res}
        \vspace{-0.5cm}
\end{table}

%



\section{Concluding Remarks}

This paper proposes a novel approach for artefact prediction in IT troubleshooting conversations that uses an ensemble of unsupervised and supervised model.
The unsupervised model leverages domain knowledge for artefact extraction without using any label data. The supervised model leverages FastText model trained on domain specific data which is fused with BERT model for improved performance with minimal label data. 
Through experimental study, we show that both the unsupervised and supervised model outperform existing state-of-the-art models on our conversation dataset. Further, the proposed ensemble of unsupervised and supervised model achieves superior performance than using either one of them individually. 
\bibliographystyle{acl_natbib}
\bibliography{triage}
\end{document}

%% file: conv_disent.tex
Conversation data from collaboration channels consists of message exchanges among SREs. These messages may consist of several different conversation threads. The conversation disentanglement module extracts messages which are part of the same thread for further analysis. \citet{LowePSCLP17} and \citet{LowePSP15} proposed a heuristic approach based on time-difference and direct message to extract threads from the Ubuntu corpus. Conversations contained in collaboration platforms, that SREs use, are multi-interlocutor in nature, whereas Ubuntu conversation is a two-way (or dyadic) conversation. SRE's collaboration platforms also have a feature that allows participants to discuss in threaded structure and these native threads can be extracted using the channel's meta-features. But since several participants may not use this feature all the time, messages may also be written outside these threads as well. These messages can be called  ~\textit{contextual messages}. This module identifies all contextual messages and merges/links them with relevant threads. Our approach extracts all native threads and extracts potential contextual messages before and after the thread. It consists of following rules to identify contextual messages:~\textit{Temporal window:} extract a set of messages within a certain temporal window as potential contextual messages, $M_{cm}$; ~\textit{User overlap:} extract a set of participant users $U_{t}$ from the thread and the set of all users $U_{c}$ from potential contextual messages. All messages from the set $M_{cm}$ written by $U_{t} {\cap}  U_{c}$ are considered part of the thread, and are merged together to form one conversation.

%% file: main.bbl
\begin{thebibliography}{15}
\expandafter\ifx\csname natexlab\endcsname\relax\def\natexlab#1{#1}\fi

\bibitem[{Ayachitula and Khandekar(2020)}]{Ayachit20}
Arun Ayachitula and Rohit Khandekar. 2020.
\newblock \href
  {https://www.linkedin.com/pulse/ai-topic-clustering-unstructured-texts-services-ayachitula/}
  {Ai for it: Topic clustering of unstructured texts in it services}.

\bibitem[{Bojanowski et~al.(2017)Bojanowski, Grave, Joulin, and
  Mikolov}]{bojanowski2017enriching}
Piotr Bojanowski, Edouard Grave, Armand Joulin, and Tomas Mikolov. 2017.
\newblock Enriching word vectors with subword information.
\newblock \emph{Transactions of the Association for Computational Linguistics},
  5:135--146.

\bibitem[{Chen et~al.(2019)Chen, Zhuo, and Wang}]{chen2019bert}
Qian Chen, Zhu Zhuo, and Wen Wang. 2019.
\newblock {BERT} for joint intent classification and slot filling.

\bibitem[{Devlin et~al.(2018)Devlin, Chang, Lee, and
  Toutanova}]{devlin2018bert}
Jacob Devlin, Ming-Wei Chang, Kenton Lee, and Kristina Toutanova. 2018.
\newblock {BERT}: {P}re-training of deep bidirectional transformers for
  language understanding.
\newblock \emph{arXiv preprint arXiv:1810.04805}.

\bibitem[{{Forsythand} and {Martell}(2007)}]{Forsythand2007}
E.~N. {Forsythand} and C.~H. {Martell}. 2007.
\newblock Lexical and discourse analysis of online chat dialog.
\newblock In \emph{International Conference on Semantic Computing (ICSC 2007)},
  pages 19--26.

\bibitem[{Gildea and Jurafsky(2002)}]{Gildea}
Daniel Gildea and Daniel Jurafsky. 2002.
\newblock \href {https://doi.org/10.1162/089120102760275983} {Automatic
  labeling of semantic roles}.
\newblock \emph{Comput. Linguist.}, 28(3):245--288.

\bibitem[{Kumar et~al.(2018)Kumar, Agarwal, Dasgupta, and Joshi}]{CRF-BILSTM}
Harshit Kumar, Arvind Agarwal, Riddhiman Dasgupta, and Sachindra Joshi. 2018.
\newblock Dialogue act sequence labeling using hierarchical encoder with crf.
\newblock In \emph{Proceedings of the AAAI Conference on Artificial
  Intelligence}, volume~32.

\bibitem[{Lowe et~al.(2015)Lowe, Pow, Serban, and Pineau}]{LowePSP15}
Ryan Lowe, Nissan Pow, Iulian Serban, and Joelle Pineau. 2015.
\newblock \href {https://doi.org/10.18653/v1/w15-4640} {The ubuntu dialogue
  corpus: {A} large dataset for research in unstructured multi-turn dialogue
  systems}.
\newblock In \emph{Proceedings of the {SIGDIAL} 2015 Conference, The 16th
  Annual Meeting of the Special Interest Group on Discourse and Dialogue, 2-4
  September 2015, Prague, Czech Republic}, pages 285--294. The Association for
  Computer Linguistics.

\bibitem[{Lowe et~al.(2017)Lowe, Pow, Serban, Charlin, Liu, and
  Pineau}]{LowePSCLP17}
Ryan~Thomas Lowe, Nissan Pow, Iulian~Vlad Serban, Laurent Charlin, Chia{-}Wei
  Liu, and Joelle Pineau. 2017.
\newblock \href {http://dad.uni-bielefeld.de/index.php/dad/article/view/3698}
  {Training end-to-end dialogue systems with the ubuntu dialogue corpus}.
\newblock \emph{Dialogue Discourse}, 8(1):31--65.

\bibitem[{Mohapatra et~al.(2018)Mohapatra, Deng, Gupta, Dasgupta, Paradkar,
  Mahindru, Rosu, Tao, and Aggarwal}]{mohapatra2018domain}
Prateeti Mohapatra, Yu~Deng, Abhirut Gupta, Gargi Dasgupta, Amit Paradkar,
  Ruchi Mahindru, Daniela Rosu, Shu Tao, and Pooja Aggarwal. 2018.
\newblock Domain knowledge driven key term extraction for it services.
\newblock In \emph{International Conference on Service-Oriented Computing},
  pages 489--504. Springer.

\bibitem[{Popov et~al.(2019)Popov, Bulatov, Polyudova, and
  Veselova}]{popov2019unsupervised}
Artem Popov, Victor Bulatov, Darya Polyudova, and Eugenia Veselova. 2019.
\newblock Unsupervised dialogue intent detection via hierarchical topic model.
\newblock In \emph{Proceedings of the International Conference on Recent
  Advances in Natural Language Processing (RANLP 2019)}, pages 932--938.

\bibitem[{Pradhan et~al.(2004)Pradhan, Ward, Hacioglu, Martin, and
  Jurafsky}]{pradhan2004shallow}
Sameer~S Pradhan, Wayne~H Ward, Kadri Hacioglu, James~H Martin, and Daniel
  Jurafsky. 2004.
\newblock Shallow semantic parsing using support vector machines.
\newblock In \emph{HLT-NAACL}, pages 233--240.

\bibitem[{Roth and Woodsend(2014)}]{roth2014composition}
Michael Roth and Kristian Woodsend. 2014.
\newblock Composition of word representations improves semantic role labelling.
\newblock In \emph{Proceedings of the 2014 Conference on Empirical Methods in
  Natural Language Processing (EMNLP)}, pages 407--413.

\bibitem[{Shrestha and McKeown(2004)}]{Mckewon2004}
Lokesh Shrestha and Kathleen McKeown. 2004.
\newblock Detection of question-answer pairs in email conversations.
\newblock In \emph{Proceedings of the 20th International Conference on
  Computational Linguistics}, COLING `04, page 889, USA. Association for
  Computational Linguistics.

\bibitem[{Vaswani et~al.(2017)Vaswani, Shazeer, Parmar, Uszkoreit, Jones,
  Gomez, Kaiser, and Polosukhin}]{vaswani2017attention}
Ashish Vaswani, Noam Shazeer, Niki Parmar, Jakob Uszkoreit, Llion Jones,
  Aidan~N Gomez, {\L}ukasz Kaiser, and Illia Polosukhin. 2017.
\newblock Attention is all you need.
\newblock In \emph{Advances in neural information processing systems}, pages
  5998--6008.

\end{thebibliography}
